\documentclass{article}
\usepackage{spconf}
\usepackage{macros_asoni}
\title{Noisy Inductive Matrix Completion Under Sparse Factor Models} 

\twoauthors
 {Akshay Soni, Troy Chevalier}
	{Yahoo! Research\\
	Sunnyvale, CA, USA\\
	{\tt \{akshaysoni,troyc\}@yahoo-inc.com}}
 {Swayambhoo Jain}
	{Department of ECE\\
	University of Minnesota -- Twin Cities\\
	{\tt jainx174@umn.edu}}

\begin{document}
\ninept
\maketitle
\begin{abstract}
Inductive Matrix Completion (IMC) is an important class of matrix completion problems that allows direct inclusion of available features to enhance estimation capabilities. These models have found applications in personalized recommendation systems, multilabel learning, dictionary learning, etc. This paper examines a general class of noisy matrix completion tasks where the underlying matrix is following an IMC model i.e., it is formed by a mixing matrix (a priori unknown) sandwiched between two known feature matrices. The mixing matrix here is assumed to be well approximated by the product of two sparse matrices---referred here to as ``sparse factor models.'' We leverage the main theorem of \cite{Soni:2016:NMC} and extend it to provide theoretical error bounds for the sparsity-regularized maximum likelihood estimators for the class of problems discussed in this paper. The main result is general in the sense that it can be used to derive error bounds for various noise models. In this paper, we instantiate our main result for the case of Gaussian noise and provide corresponding error bounds in terms of squared loss.
%
%
%
%
%
%
\end{abstract}
\section{Introduction}
In recent years, matrix completion has been a venue for constant research---both from theoretical as well as algorithmic front. The applications include collaborative filtering, multiclass learning, dimensionality reduction etc. In most general terms, the aim of these problems is to estimate the elements of a matrix $\bX^* \in \RR^{n_1 \times n_2}$ from noisy observations collected on only a subset of its locations. Of course, such estimation problems are ill-posed without further assumptions, since the value of $\bX^*$ at the unobserved locations can be arbitrary. A common approach is to augment the inference method with the assumption that the underlying matrix to be estimated exhibits some form of intrinsic low-dimensional structure, making, in some cases, the problem feasible. For example, the assumption of low-rank on the target matrix has been the main focus of research for these problems. Even though the rank constrained problem is non-convex and difficult to analyze, convex relaxations to these problems have been well studied in noiseless \cite{Candes:09:Exact}, additive uncertainty \cite{Keshavan:10:Noisy}, and even in settings where observations are nonlinear (e.g., one-bit quantized \cite{Davenport:2014:onebit}). However, due to the computational complexity reasons, the most popular algorithms to solve these problems are variants of alternating minimization \cite{Soni:2016:NMC}; the low-rank structure is imposed explicitly via bilinear models i.e., $\bX^*$ is assumed to be a product of two factor-matrices as 
%
%
%
%
%
%
\begin{equation}\label{eqn:sparse-factor}
\bX^* = \bD^* \bA^*,
\end{equation}
where $\bD^* \in \RR^{n_1 \times r}$, $\bA^* \in \RR^{r \times n_2}$ and $r \ll {\rm min}(n_1, n_2)$. Slight modifications of this formulation also allows for explicit control and facilitates imposing additional constraints like positivity \cite{Soni:2014:Poisson}, sparsity \cite{Soni:2014:NMC}, or even tree-sparsity \cite{Soni:2014:Tree} on the factors. 

In terms of the well-known user-item recommendation problem \cite{Koren:09}: $\bX^*$ pertains to the rating matrix, $\bD^*$ and $\bA^*$ denote the matrices of user and item latent-factors, respectively. Apart from the rating information, generally, we also have other user and item side-features (like age, gender, etc) that can be used to further improve the quality of estimation. These features can be included via a Bayesian framework \cite{Ma:2008:BMF,Porteous:2010:Bayesian} or by directly changing the model as in \emph{Inductive Matrix Completion} (IMC) \cite{Jain:2013:IMC,Xu:2013:IMC} where we have the following model assumption on $\bX^*$:
\begin{equation}\label{eqn:imc}
\bX^* = \bA\bW^*\bB = \bA \bP^* \bQ^* \bB,
\end{equation}
where $\bA \in \RR^{n_1 \times r_1}$, $\bB \in \RR^{r_2 \times n_2}$ are known matrices, for example, user and item feature matrices, $\bW^* \in \RR^{r_1 \times r_2}$ is a mixing matrix whose low-rank structure is explicitly imposed via a bilinear model $\bW^* = \bP^*\bQ^*$, $\bP^* \in \RR^{r_1 \times r}$ and $\bQ^* \in \RR^{r \times r_2}$. 

The goal of this paper is to provide estimation error bounds for matrix completion under inductive setting, as in \eqref{eqn:imc}, by using the framework of regularized maximum likelihood estimation \cite{Soni:2016:NMC}. For generality, we derive these bounds under the assumption that both $\bP^*$ and $\bQ^*$ can be sparse, and noise can follow any generic distribution. This kind of sparse-structure arises in recommendation problems where the desire is to have sparse latent vectors due to requirement of low computational and storage complexity \cite{Acar:2012:sparsefactors}, and variant of dictionary learning problems where apart from the coefficient matrix being sparse, dictionary itself is modeled to be sparse \cite{Rubinstein:2010:doublesparse}. 
%
%
%
%
%
%
%
%
%

%
%
%

\subsection{Related Works}
Our inference tasks here essentially entails learning two structured factors in a bilinear model. With a few notable exceptions (e.g., low-rank matrices and certain non-negative matrices \cite{Arora:12,Esser:12}), the non-convex bilinear models studied in this paper are difficult to analyze. Several recent efforts in the dictionary learning literature have established theoretical guarantees on identifiability, as well as algorithmic analyses of a number of optimization approaches \cite{Haupt:2014:onebit,Gribonval:10:Identification,Aharon:06:Uniqueness,Spielman:13:Exact}.
The most closely related to our work are \cite{Soni:2016:NMC,Soni:2014:NMC,Haupt:2014:onebit}, where authors provide a framework for deriving estimation error bounds for the models of form \eqref{eqn:sparse-factor} under various noise distributions and sparsity structure on $\bA^*$ by using a regularized maximum-likelihood approach based on the ideas from \cite{Barron:08:MDL}. The Overall theme of the estimation bounds obtained using this framework is that the error of their estimate, ${\hat \bX}$ of $\bX^*$, decays as the ratio of number of degrees of freedom to estimate over the number of observations.
%
%
%
%
%
%
%
%
%

The idea of including side features in the factorization models to improve recommendations is considered using Bayesian formulations in \cite{Agarwal:2010:flda}  or via directly incorporating these features in a generative model in IMC \cite{Xu:2013:IMC}, Factorization Machines \cite{Rendle:2010:FM}. The statistical performance of IMC has been theoretically analyzed via convex relaxations based on nuclear-norm in noiseless \cite{Xu:2013:IMC} and noisy settings \cite{Chiang:2015:NIMC}. From the algorithmic perspective, provable alternating-minimization based approach with bilinear models has been recently proposed and analyzed \cite{Jain:2013:IMC}. Our results in this paper provide additional insights into these problems under a general setting of noise and sparsity of the bilinear factors. It is to be noted here that the aim of all these related works is to accurately estimate the mixing matrix $\bW^*$; we provide error bounds for the estimation of $\bX^*$. 
%
%
%

Many recent works have used IMC framework for large scale recommendations \cite{Shin:2015:tumblr} or extended IMC to automatically construct better features \cite{Si:2016:goaldirec}. 
%
%

\subsection{Specific Contributions}
In this paper we provide estimation error guarantees for matrix completion using regularized maximum likelihood estimator under the model described in \eqref{eqn:imc} for the cases where the factor matrices are dense (either one or both), or when $\bW^*$ is itself sparse. In order to  arrive at these bounds, we extend the fundamental theorem proposed in \cite{Soni:2016:NMC} to the inductive matrix completion setting studied here. This extension is detailed in Theorem \ref{thm:main}. The main utility of this theorem is that it can provide estimation error guarantees for a variety of noise distributions; for example, Gaussian, Poisson, Laplace, and even under the extreme quantized setting of 1-bit observations. Here we  instantiate it for the Gaussian noise and obtain corresponding error guarantees.  

\subsection{Preliminaries}
To set the stage for the statement of our main result, we recall several information-theoretic preliminaries.  When $p(Y)$ and $q(Y)$ denote the pdf (or pmf) of a real-valued random variable $Y$, the Kullback-Leibler divergence (or KL divergence) of $q$ from $p$ is denoted $\D(p\|q)$ and given by
\begin{equation*}
\D(p\|q) = 
\E_{p}\left[\log\frac{p(Y)}{q(Y)}\right]
\end{equation*}
where the logarithm is taken to be the natural log.  By definition, $\D(p\|q)$ is finite only if the support of $p$ is contained in the support of $q$.  Further, the KL divergence satisfies $\D(p\|q) \geq 0$ and $\D(p\|q) = 0$ when $p(Y) = q(Y)$.   We also use the Hellinger affinity denoted by $\A(p,q)$ and given by
\begin{equation*}
\A(p,q) = \E_{p}\left[\sqrt{\frac{q(Y)}{p(Y)}}\right] = \E_{q}\left[\sqrt{\frac{p(Y)}{q(Y)}}\right]
\end{equation*}
It is known that $0 \leq \A(p,q) \leq 1$.  When $p$ and $q$ are parameterized by elements $X_{i,j}$ and $\widetilde{X}_{i,j}$ of matrices $\bX$ and $\widetilde{\bX}$, respectively, so that $p(Y_{i,j}) = p_{X_{i,j}}(Y_{i,j})$ and $q(Y_{i,j}) = q_{\widetilde{X}_{i,j}}(Y_{i,j})$, we use the shorthand notation $\D(p_{\bX}\|q_{\widetilde{\bX}}) \triangleq \sum_{i,j} \D(p_{X_{i,j}}\|q_{\widetilde{X}_{i,j}})$ and $\A(p_{\bX},q_{\widetilde{\bX}}) \triangleq \prod_{i,j} \A(p_{X_{i,j}},q_{\widetilde{X}_{i,j}})$. Finally, for a matrix $\bM$ we denote by $\|\bM\|_0$ its number of nonzero elements, and $\|\bM\|_{\rm max}$ the magnitude of its largest element (in absolute value). Also, $(a_1 \vee a_2) = {\rm max}(a_1, a_2)$.

\section{Problem Statement and Recovery Result}\label{problem}
Our focus here is to estimate a matrix $\bX^*$ that admits a model as in \eqref{eqn:imc} from noisy observations $\bY_{\cal S} = \{Y_{i,j}\}_{(i,j) \in {\cal S}}$ collected at a subset ${\cal S} \subset [n_1] \times [n_2]$ of its locations. Sampling locations are randomly chosen in the sense that for an integer $m$ satisfying $4 \leq m \leq n_1n_2$ and $\gamma = m(n_1 n_2)^{-1}$, each location is observed independently via a Bernoulli($\gamma$) model. Going forward, we assume the following bounds on the participating matrices:
\begin{eqnarray*}
\|\bX^*\|_{\rm max} \leq {\rm X}_{\rm max}/2, ~\|\bP^*\|_{\rm max} \leq 1,~ \|\bQ^*\|_{\rm max} \leq {\rm Q}_{\rm max}, \\
~\|\bA\|_{\rm max} \leq {\rm A}_{\rm max},~\text{and}~\|\bB\|_{\rm max} \leq {\rm B}_{\rm max},
\end{eqnarray*}
for some positive constants ${\rm Q}_{\rm max}$, ${\rm A}_{\rm max}$, and ${\rm B}_{\rm max}$.

Given ${\cal S}$ and $\bY_{\cal S}$, we write the joint pdf (or pmf) of the observations as
\begin{equation}\label{eqn:obsmodel}
p_{\bX^*_{\cS}}(\bY_{\cS}) \triangleq \prod_{(i,j)\in\cS} p_{X^*_{i,j}}(Y_{i,j}),
\end{equation} 
where $p_{X^*_{i,j}}(Y_{i,j})$ denotes the corresponding scalar pdf (or pmf), and we use the shorthand $\bX^*_{\cS}$ to denote the collection of elements of $\bX^*$ indexed by $(i,j)\in\cS$.  Then our task may be described succinctly as follows:  given $\cS$ and corresponding noisy observations $\bY_{\cS}$ of $\bX^*$ distributed according to \eqref{eqn:obsmodel}, our goal is to estimate $\bX^*$ under the assumption that it admits a model as in \eqref{eqn:imc}. 

Our approach will be to estimate $\bX^*$ via sparsity-penalized maximum likelihood; we consider estimates of the form
\begin{equation}\label{eqn:xhatsparse}
\widehat{\bX} = \arg\min_{\bX= \bA\bP\bQ\bB \in \cX} \ \left\{-\log p_{\bX_{\cS}}(\bY_{\cS}) + \lambda_{\rm P} \|\bP\|_{0} + \lambda_{\rm Q} \|\bQ\|_{0} \right\},
\end{equation}
where $\{\lambda_{\rm P}, \lambda_{\rm Q}\} >0$ are user-specified regularization parameter, $\bX_{\cS}$ is shorthand for the collection $\{X_{i,j}\}_{(i,j)\in\cS}$ of entries of $\bX$ indexed by $\cS$, and $\cX$ is an appropriately constructed class of candidate estimates.  To facilitate our analysis here, we take $\cX$ to be a countable class of estimates constructed as follows: for a specified $\beta \geq 1$, we set $L_{\rm lev} = 2^{\lceil\log_2(n_1 \vee n_2)^\beta \rceil}$ and construct $\cal P$ to be the set of all matrices $\bP \in \RR^{r_1 \times r}$ whose elements are either $0$, or are discretized to one of $L_{\rm lev}$ uniformly spaced levels in the range $[-1, 1]$ and ${\cal Q}$ to be the set of all matrices $\bQ \in \RR^{r \times r_2}$ whose elements either take the value zero, or are discretized to one of $L_{\rm lev}$ uniformly-spaced levels in the range $[-{\rm Q}_{\rm max}, {\rm Q}_{\rm max}]$. Then, we let
\begin{equation}
{\cal X} \triangleq \{ \bX = \bA\bP\bQ\bB~:~ \bP \in {\cal P},~ \bQ \in {\cal Q}, ~\|\bX\|_{\rm max} \leq {\rm X}_{\rm max}\}.
\end{equation}

Our main result establishes error bounds under general noise or corruption models. We state the result here as a theorem; its proof appears in Section \ref{proofs}.

\begin{thmi}\label{thm:main}
Let $\cS$, $\gamma$, $m$, $\beta$, $\cX$, and $\bY_{\cS}$ be as defined above and $\cD$ is any constant satisfying
\begin{equation*}
\cD \geq \max_{\bX\in\cX} \max_{i,j}  \ D(p_{X^*_{i,j}}\|p_{X_{i,j}}).
\end{equation*} 
Then for any
\begin{eqnarray} \label{eqn:reg}
\lambda_{\rm P} \geq 2 \left(1+\frac{2\cD}{3}\right) \left(4\log(r_1) + \beta \log(n_1 \vee n_2)\right) \\
\lambda_{\rm Q} \geq 2 \left(1+\frac{2\cD}{3}\right) \left(4\log(r_2) + \beta \log(n_1 \vee n_2)\right)
\end{eqnarray}
the complexity penalized maximum likelihood estimator \eqref{eqn:xhatsparse}
satisfies the (normalized, per-element) error bound
\begin{eqnarray*}
\lefteqn{\frac{\E_{\cS,\bY_{\cS}}\left[-2\log \A(p_{\widehat{\bX}^{\lambda}}, p_{\bX^*})\right]}{n_1 n_2} \leq \frac{8 \cD \log m}{m} +}\\
&&3 \cdot \min_{\bX\in\cX} \left\{ \frac{\D(p_{\bX^*}\|p_{\bX})}{n_1 n_2} + \right.\\
&&\left. \left((\lambda_{\rm P} + \lambda_{\rm Q}) + \frac{4\cD (\beta + 4) \log(n_1\vee n_2)}{3}\right)\left(\frac{\|\bP\|_0 + \|\bQ\|_0}{m}\right)\right\}.
\end{eqnarray*} 
\end{thmi} 

Before proceeding to specific case of Gaussian noise, we note a few salient points about this result. First, as alluded above, our results is general and can be used to analyze the error performance under a variety of noise models. Specialization to a given noise model requires us to compute the upper bounds for KL divergences in terms of problem parameters, and the lower bound of negative log Hellinger affinities in terms of the required error function. Second, the above result is kind of an \emph{oracle} bound, in that it is specified in terms of a minimum over $\bX \in \cX$. We can construct a valid upper bound by evaluating this for any $\bX \in \cX$. We construct a specific $\bX \in \cX$ that is close to $\bX^*$, and evaluate the oracle bound for the constructed point to get a non-trivial upper bound.
%
%
%

Finally, we note that the optimization problem \eqref{eqn:xhatsparse} is non-convex, as often is the case, not only because of $\ell_0$ penalty, or discretized space ($\cX$) used for proofs in this paper, but due the fundamental challenge posed by bilinear models. 
%
%
%

\subsection{Implications for Gaussian Noise} \label{implications}
Below we provide scaling behavior of error bound with problem parameters for Gaussian noise with known variance by instantiating Theorem \ref{thm:main}. The joint likelihood of observations can be written as
\begin{equation}\label{eqn:likGauss}
p_{\bX^*_{S}}(\bY_{S}) = \frac{1}{(2\pi\sigma^2)^{|S|/2}}\exp\left(-\frac{1}{2 \sigma^2} \ \|\bY_{S} - \bX^*_{S}\|_F^2\right),
\end{equation}
where we have used the shorthand notation $\|\bY_{S} - \bX^*_{S}\|_F^2 \triangleq \sum_{(i,j)\in S} (Y_{i,j}-X^*_{i,j})^2$.  To that end, we fix $\beta$ as 
\begin{equation}\label{beta}
\beta = {\rm max}\left\{1, 1 + \frac{\log (6 \sqrt{m} \cdot (r \cdot r_1 \cdot r_2) \cdot \Amax \Bmax \Qmax / \Xmax)}{\log(n_1 \vee n_2)}\right\}
\end{equation}
for describing the number of discretization levels in the elements of each factor matrix and regularization parameters $(\lambda_{\rm P}, \lambda_{\rm Q})$ equal to right hand side of \eqref{eqn:reg}. In this setting  we have the following result; its proof appears in Section~\ref{gaussproof}.

\begin{cori}\label{cor:Gauss}
Let $\beta$, $\lambda_{\rm P}$, and $\lambda_{\rm Q}$ be as defined above with $\cD = 2\Xmax^2/\sigma^2$. The estimate $\widehat{\bX}$ obtained via \eqref{eqn:xhatsparse} satisfies
\begin{eqnarray} \label{eqn:gausssparse}
\lefteqn{\frac{\E_{\cS,\bY_{\cS}}\left[\|\bX^*-\widehat{\bX}\|_F^2\right]}{n_1 n_2} =}\\ 
\nonumber & &{\cal O}\left((\sigma^2 + \Xmax^2)\left(\frac{\|\bP^*\|_0 + \|\bQ^*\|_0}{m}\right)\log(n_1\vee n_2)\right).
\end{eqnarray}
\end{cori}

A few comments are in order regarding these error guarantees. We may interpret
the quantity $\|\bP^*\|_0 + \|\bQ^*\|_0$ as the number of degrees of freedom
to be estimated, and in this sense we see that the error
rate of the penalized maximum likelihood estimator exhibits characteristics
of the well-known parametric rate (modulo the logarithmic
factor). If $\bP^*$ and $\bQ^*$ are instead dense, then we get the familiar error bound that decays as $(r_1 + r_2)r/m$ with $m$. Compared this with the case where matrix $\bX^*$ is just low-rank (as in formal matrix completion tasks), the error rates are of the form $(n_1 + n_2)r/m$ \cite{Soni:2016:NMC}. Clearly, the IMC model has a better error performance if $r_2 \ll n_2$ or even better if the factor matrices are sparse.

The presence of the factor ${\rm X}_{\rm max}^2$ is akin to the incoherence conditions prevalent in the matrix-completion literature. In essence, it pertains to the spikiness of the underlying matrix we are trying to estimate \cite{Negahban:12}.

\section{Proofs}\label{proofs}
\subsection{Proof of Theorem \ref{thm:main}}
We utilize the following Lemma from \cite{Soni:2016:NMC} and apply it to our problem. 
\begin{lemmai}\label{lem:main}
Let $\bX^*$ be an $n_1\times n_2$ matrix whose elements we aim to estimate, and let $\cX$ be a countable collection of candidate reconstructions $\bX$ of $\bX^*$, each with corresponding penalty $\pen(\bX)\geq 1$, so that the collection of penalties satisfies the summability condition $\sum_{\bX\in\cX} 2^{-\pen(\bX)} \leq 1$. Given $m$, $\gamma$, $\cS$, $\bY_{\cS}$, and $\cD$ as explained earlier, 
we have that for any  $\xi \geq \left(1 + \frac{2\cD}{3}\right) \cdot 2\log 2$, the complexity penalized maximum likelihood estimator
\begin{eqnarray*}
\widehat{\bX}^{\xi} =  \arg\min_{\bX\in \cX}\left\{-\log p_{\bX_{\cS}}(\bY_{\cS}) + \xi \cdot \pen(\bX)  \right\},
\end{eqnarray*}
satisfies the (normalized, per-element) error bound
\begin{eqnarray*}
\frac{\E_{\cS,\bY_{\cS}}\left[-2\log \A(p_{\widehat{\bX}^{\xi}}, p_{\bX^*})\right]}{n_1 n_2} \leq \frac{8 \cD \log m}{m} + \\
3 \cdot \min_{\bX\in\cX} \left\{ \frac{\D(p_{\bX^*}\|p_{\bX})}{n_1 n_2} + \left(\xi + \frac{4\cD\log 2}{3}\right)\frac{\pen(\bX)}{m}\right\},
\end{eqnarray*}
where, as denoted, the expectation is with respect to the joint distribution of $\cS$ and $\bY_{\cS}$.
\end{lemmai}

Now, consider any discretized matrix factors $\bP \in {\cal P}$ and $\bQ \in {\cal Q}$, as described in Section \ref{problem}. For $L_{\rm loc}^{\cal P} \triangleq 2^{\lceil \log_2 (r_1\cdot r)\rceil}$, we encode each nonzero element of $\bP$ using $\log_2 L_{\rm loc}^{\cal P}$ bits to denote its location and $\log_2 L_{\rm lev}$ bits for its amplitude. We can then encode $\bP$ with $\|\bP\|_0$ nonzero entries by using $\|\bP\|_0(\log_2 L_{\rm loc}^{\cal P} + \log_2 L_{\rm lev})$ bits. Similarly, we can encode $\bQ$ using $\|\bQ\|_0(\log_2 L_{\rm loc}^{\cal Q} + \log_2 L_{\rm lev})$ where $L_{\rm loc}^{\cal Q} \triangleq 2^{\lceil \log_2 (r_2\cdot r)\rceil}$. Now, we let ${\cal X}'$ be set of all such $\bX = \bA \bP \bQ \bB$, and let the binary code for each $\bX$ be the concatenation of the binary code for $\bP$ followed by the binary code for $\bQ$. It follows that we may assign penalties $\pen(\bX)$ to all $\bX \in {\cal X}'$ whose length satisfy
\begin{equation}
\pen(\bX) = \|\bP\|_0(\log_2 (L_{\rm loc}^{\cal P}\cdot L_{\rm lev})) + \|\bQ\|_0(\log_2 (L_{\rm loc}^{\cal Q} \cdot L_{\rm lev})),
\end{equation}
given that $\bA$ and $\bB$ are known matrices.

It is easy to see that such codes are (by construction) uniquely decodable, so we have that $\sum_{\bX\in\cX'} 2^{-\pen(\bX)} \leq 1$ by the well-known Kraft-McMillan Inequality \cite{Cover:12}. Now let $\cX$ be any subset of $\cX'$. For randomly subsampled and noisy observations $\bY_{\cS}$ our estimates take the form
\begin{eqnarray*}
\nonumber \widehat{\bX}^{\xi} 
&=&  \arg\min_{\bX= \bA \bP\bQ \bB \in \cX}\left\{-\log p_{\bX_{\cS}}(\bY_{\cS}) + \xi \cdot  \pen(\bX) \right\}.
\end{eqnarray*}
Further, using Lemma \ref{lem:main} we have
\begin{eqnarray*}
\lefteqn{\frac{\E_{\cS,\bY_{\cS}}\left[-2\log \A(p_{\widehat{\bX}^{\xi}}, p_{\bX^*})\right]}{n_1 n_2} \leq \frac{8 \cD \log m}{m} +}\\
&& \hspace{-2em}3 \cdot \min_{\bX\in\cX} \left\{ \frac{\D(p_{\bX^*}\|p_{\bX})}{n_1 n_2} + \right. \\
&& \hspace{-2em} \left. \left(\xi + \frac{4\cD\log 2}{3}\right)(\log_2 L_{\rm loc} + \log_2 L_{\rm lev})\left(\frac{\|\bP\|_0 + \|\bQ\|_0}{m}\right)\right\},
\end{eqnarray*}
where $\log_2 L_{\rm loc} = \left(\log_2 L_{\rm loc}^{\cal P} \vee  ~\log_2 L_{\rm loc}^{\cal Q}\right)$. Finally, letting 
\begin{eqnarray*}
\lambda_{\rm P} &=& \xi \cdot  (\log_2 L_{\rm loc}^{\cal P} + \log_2 L_{\rm lev})\\
\lambda_{\rm Q} &=& \xi \cdot  (\log_2 L_{\rm loc}^{\cal Q} + \log_2 L_{\rm lev})
\end{eqnarray*}
and using the fact that
\begin{eqnarray*}
\log_2 L_{\rm loc}^{\cal P} + \log_2 L_{\rm lev}  &\leq& 8 \log(r_1) + 2\beta \log(n_1 \vee n_2) \\
\log_2 L_{\rm loc}^{\cal Q} + \log_2 L_{\rm lev}  &\leq& 8 \log(r_2) + 2\beta \log(n_1 \vee n_2)  \\
\log_2 L_{\rm loc} + \log_2 L_{\rm lev}  &\leq& (\beta + 4)  \cdot \log(n_1\vee n_2) \cdot 2\log_2(e)
\end{eqnarray*}
which follows by our selection of $L_{\rm lev}$ and $L_{\rm loc}$ and the fact that $r_1 < n_1$, $r_2 < n_2$; it follows (after some straightforward simplification) that ($\lambda_{\rm P}, \lambda_{\rm Q}$) follow \eqref{eqn:reg}, and the estimate \eqref{eqn:xhatsparse} satisfies 
\begin{eqnarray*}
\lefteqn{\frac{\E_{\cS,\bY_{\cS}}\left[-2\log \A(p_{\widehat{\bX}}, p_{\bX^*})\right]}{n_1 n_2} \leq \frac{8 \cD \log m}{m} +}\\
&&3 \cdot \min_{\bX\in\cX} \left\{ \frac{\D(p_{\bX^*}\|p_{\bX})}{n_1 n_2} + \right.\\
&&\left. \left((\lambda_{\rm P} + \lambda_{\rm Q}) + \frac{4\cD (\beta + 4) \log(n_1\vee n_2)}{3}\right)\left(\frac{\|\bP\|_0 + \|\bQ\|_0}{m}\right)\right\},
\end{eqnarray*} 
as claimed.

\subsection{Proof of Corollary~\ref{cor:Gauss}} \label{gaussproof}
\begin{proof}
For $\bX^*$ as specified and any $\bX\in\cX$, using the model \eqref{eqn:likGauss} we have
\begin{equation*}
\D(p_{X_{i,j}^*}\|p_{X_{i,j}}) =  \frac{(X_{i,j}^*-X_{i,j})^2}{2\sigma^2}
\end{equation*}
for any fixed $(i,j)\in S$. It follows that $\D(p_{\bX^*}\|p_{\bX})=\|\bX^*-\bX\|_F^2/2\sigma^2$, and using the fact that  the amplitudes of entries of $\bX^*$ and all $\bX\in\cX$ are no larger than $\Xmax$, it is clear that we may choose $\cD = 2 \Xmax^2/\sigma^2$.  Further, for any $\bX\in\cX$ and any fixed $(i,j)\in\cS$ it is easy to show that in this case
\begin{equation*}
-2\log \A(p_{X_{i,j}}, p_{X_{i,j}^*}) =  \frac{(X_{i,j}^*-X_{i,j})^2}{4\sigma^2},
\end{equation*}
so that $-2\log \A(p_{\bX}, p_{\bX^*})=\|\bX^*-\bX\|_F^2/4\sigma^2$.  It follows that 
\begin{equation*}
\E_{\cS, \bY_{\cS}} \left[ -2 \log \A(p_{\widehat{\bX}},p_{\bX^*}) \right] = \frac{\E_{\cS, \bY_{\cS}} \left[ \|\bX^*-\widehat{\bX}\|_F^2 \right]}{4\sigma^2}.
\end{equation*}
Incorporating this into Theorem~\ref{thm:main}, we obtain that for regularization parameters as in \eqref{eqn:reg} with $\cD$ substituted, the sparsity penalized maximum-likelihood estimate satisfies the per-element mean-square error bound
\begin{eqnarray*}
\nonumber  \lefteqn{\frac{\E_{\cS,\bY_{\cS}}\left[\|\bX^*-\widehat{\bX}\|_F^2\right]}{n_1 n_2} \leq  \frac{64 \Xmax^2 \log m}{m}}\\ 
&& \hspace{-2em}\mbox{} + 6 \cdot \min_{\bX\in\cX} \left\{ \frac{\|\bX^*-\bX\|_F^2}{n_1 n_2}  ~+ \right.\\ 
&& \hspace{-2em} \left. \left(2\sigma^2 (\lambda_{\rm P} + \lambda_{\rm Q}) + \frac{16\Xmax^2 (\beta+4)\log (n_1\vee n_2)}{3}\right)\right.\\
&& \hspace{8em} \left. \left(\frac{\|\bP\|_0 + \|\bQ\|_0}{m}\right)\right\}.
\end{eqnarray*}

We now establish the error bound for the case where $\bP^*$ and $\bQ^*$ are sparse with corresponding cardinality $\|\bP^*\|_0$ and $\|\bQ^*\|_0$. Consider a candidate reconstruction of the form $\bX^*_{q} = \bA \bP^*_{q}\bQ^*_{q} \bB$, where the elements of $\bP^*_{q}$ are the closest discretized surrogates of the \emph{nonzero} entries of $\bP^{*}$, and the entries of and $\bQ^*_{q}$  are the closest discretized surrogates of the \emph{nonzero} entries of $\bQ^{*}$. Denote $\bP^*_{q} = \bP^{*} + \triangle_{\bP^*}$ and $\bQ^*_{q} = \bQ^{*} + \triangle_{\bQ^*}$. Then it is easy to see that
\begin{equation*}
\bX^*_{q} - \bX^{*} = \bA(\bP^{*}\triangle_{\bQ^*}  +  \triangle_{\bP^*}\bQ^{*}  +  \triangle_{\bP^*}\triangle_{\bQ^*})\bB.
\end{equation*}
Given the range limits on allowable $\bP$ and $\bQ$ and that each range is quantized to $L_{\rm lev}$ levels, we have that $\|\triangle_{\bP^*}\|_{\rm max} \leq 1/(L_{\rm lev} -1)$ and $\|\triangle_{\bQ^*}\|_{\rm max} \leq {\rm Q}_{\rm max}/(L_{\rm lev} -1)$. Now, we can obtain a bound on the magnitudes of the elements of $\bX^*_{q} - \bX^{*}$ that hold uniformly over all $i,j$, as follows
\begin{eqnarray}\label{eqn:quantmax}
\lefteqn{\nonumber \|\bX^*_{q} - \bX^{*}\|_{\rm max}}\\ 
\nonumber &=& \max_{i,j} |(\bA(\bP^{*}\triangle_{\bQ^*}  +  \triangle_{\bP^*}\bQ^{*}  +  \triangle_{\bP^*}\triangle_{\bQ^*})\bB)_{i,j}|\\
&\leq& \frac{6 \cdot r \cdot r_1 \cdot r_2 \cdot \Amax \Bmax \Qmax }{L_{\rm lev}},
\end{eqnarray}
where the inequality follows from a application of triangle inequality followed by the bounds on $\|\triangle_{\bP^*}\|_{\rm max}$ and $\|\triangle_{\bQ^*}\|_{\rm max}$ and the entry-wise bounds on elements of allowable $\bP$ and $\bQ$, and because $L_{\rm lev} \geq 2$.  Now, it is straight-forward to show that our choice of $\beta$ in \eqref{beta} implies 
\begin{equation}
L_{\rm lev} \geq \frac{12 \cdot r \cdot r_1 \cdot r_2 \cdot \Amax \Bmax \Qmax}{\Xmax},
\end{equation}
so each entry of $\bX^*_{q} - \bX^{*}$ is bounded in magnitude by $\Xmax/2$.  It follows that each element of the candidate $\bX^*_{q}$ constructed above is bounded in magnitude by $\Xmax$, so $\bX^*_q$ is indeed a valid element of the set $\cX$.

Further, the approximation error analysis above also implies directly that 
\begin{eqnarray*}
\nonumber \frac{\|\bX^{*} - \bX^*_{q}\|_{F}^{2}}{n_1n_2} &=& \frac{1}{n_1 n_2} \sum_{i \in [n_{1}], j \in [n_{2}]} (\bA\bP^*_{q}\bQ^*_{q}\bB- \bA \bP^{*}\bQ^{*} \bB)^{2}_{i,j} \\
\nonumber &\leq& \frac{36 \cdot (r \cdot r_1 \cdot r_2)^{2} (\Amax \Bmax \Qmax)^2}{L_{\rm lev}^{2}}\\
&\leq& \frac{\Xmax^2}{m} , 
\end{eqnarray*}
where the last line follows from the fact that our specific choice of $\beta$ in \eqref{beta} also implies 
\begin{equation}
L_{\rm lev} \geq \frac{6  \sqrt{m} \cdot  (r \cdot r_1 \cdot r_2) \cdot \Amax \Bmax \Qmax}{\Xmax}.
\end{equation}
Now, evaluating the oracle term at the candidate $\bX^*_q=\bA \bP^*_q\bQ^*_q \bB$, and using the fact that $\|\bP^*_q\|_0 = \|\bP^*\|_0$ and $\|\bQ^*_q\|_0 = \|\bQ^*\|_0$, we have
\begin{eqnarray*}
& &\frac{\E_{\cS,\bY_{\cS}}\left[\|\bX^*-\widehat{\bX}\|_F^2\right]}{n_1 n_2} \leq \frac{70 \Xmax^2 \log m}{m} +\\
&& 48 (\sigma^2 + 2\Xmax^2) (\beta+4)\log (n_1\vee n_2)\left(\frac{\|\bP^*\|_0 + \|\bQ^*\|_0}{m}\right).
\end{eqnarray*} 

\end{proof}

\newpage
\bibliographystyle{IEEEbib}
\bibliography{NIMC} 

\begin{thebibliography}{10}

\bibitem{Soni:2016:NMC}
A.~Soni, S.~Jain, J.~Haupt, and S.~Gonella,
\newblock ``Noisy matrix completion under sparse factor models,''
\newblock {\em IEEE Transactions on Information Theory}, vol. 62, no. 6, pp.
  3636--3661, June 2016.

\bibitem{Candes:09:Exact}
E.~J. Cand{\`e}s and B.~Recht,
\newblock ``Exact matrix completion via convex optimization,''
\newblock {\em Foundations of Computational Mathematics}, vol. 9, no. 6, pp.
  717--772, 2009.

\bibitem{Keshavan:10:Noisy}
R.~H. Keshavan, A.~Montanari, and S.~Oh,
\newblock ``Matrix completion from noisy entries,''
\newblock {\em Journal of Machine Learning Research}, vol. 11, pp. 2057--2078,
  2010.

\bibitem{Davenport:2014:onebit}
M.~A. Davenport, Y.~Plan, E.~van~den Berg, and M.~Wootters,
\newblock ``1-bit matrix completion,''
\newblock {\em Information and Inference}, vol. 3, no. 3, pp. 189--223, 2014.

\bibitem{Soni:2014:Poisson}
A.~Soni and J.~Haupt,
\newblock ``Estimation error guarantees for poisson denoising with sparse and
  structured dictionary models,''
\newblock in {\em 2014 IEEE International Symposium on Information Theory},
  June 2014, pp. 2002--2006.

\bibitem{Soni:2014:NMC}
A.~Soni, S.~Jain, J.~Haupt, and S.~Gonella,
\newblock ``Error bounds for maximum likelihood matrix completion under sparse
  factor models,''
\newblock in {\em Signal and Information Processing (GlobalSIP), 2014 IEEE
  Global Conference on}, Dec 2014, pp. 399--403.

\bibitem{Soni:2014:Tree}
A.~Soni and J.~Haupt,
\newblock ``On the fundamental limits of recovering tree sparse vectors from
  noisy linear measurements,''
\newblock {\em IEEE Transactions on Information Theory}, vol. 60, no. 1, pp.
  133--149, Jan 2014.

\bibitem{Koren:09}
Y.~Koren, R.~Bell, and C.~Volinsky,
\newblock ``Matrix factorization techniques for recommender systems,''
\newblock {\em Computer}, vol. 42, no. 8, pp. 30--37, 2009.

\bibitem{Ma:2008:BMF}
Hao Ma, Haixuan Yang, Michael~R Lyu, and Irwin King,
\newblock ``Sorec: social recommendation using probabilistic matrix
  factorization,''
\newblock in {\em Proceedings of the 17th ACM conference on Information and
  knowledge management}. ACM, 2008, pp. 931--940.

\bibitem{Porteous:2010:Bayesian}
Ian Porteous, Arthur~U Asuncion, and Max Welling,
\newblock ``Bayesian matrix factorization with side information and dirichlet
  process mixtures.,''
\newblock in {\em AAAI}, 2010.

\bibitem{Jain:2013:IMC}
Prateek Jain and Inderjit~S Dhillon,
\newblock ``Provable inductive matrix completion,''
\newblock {\em arXiv preprint arXiv:1306.0626}, 2013.

\bibitem{Xu:2013:IMC}
Miao Xu, Rong Jin, and Zhi-Hua Zhou,
\newblock ``Speedup matrix completion with side information: Application to
  multi-label learning,''
\newblock in {\em Advances in Neural Information Processing Systems}, 2013, pp.
  2301--2309.

\bibitem{Acar:2012:sparsefactors}
E.~Acar, G.~Gürdeniz, M.~A. Rasmussen, D.~Rago, L.~O. Dragsted, and R.~Bro,
\newblock ``Coupled matrix factorization with sparse factors to identify
  potential biomarkers in metabolomics,''
\newblock in {\em 2012 IEEE 12th International Conference on Data Mining
  Workshops}, Dec 2012, pp. 1--8.

\bibitem{Rubinstein:2010:doublesparse}
R.~Rubinstein, M.~Zibulevsky, and M.~Elad,
\newblock ``Double sparsity: Learning sparse dictionaries for sparse signal
  approximation,''
\newblock {\em IEEE Transactions on Signal Processing}, vol. 58, no. 3, pp.
  1553--1564, March 2010.

\bibitem{Arora:12}
Sanjeev Arora, Rong Ge, Ravindran Kannan, and Ankur Moitra,
\newblock ``Computing a nonnegative matrix factorization--provably,''
\newblock {\em Proceedings of the forty-fourth annual ACM symposium on Theory
  of computing}, pp. 145--162, 2012.

\bibitem{Esser:12}
E.~Esser, M.~Moller, S.~Osher, G.~Sapiro, and J.~Xin,
\newblock ``A convex model for nonnegative matrix factorization and
  dimensionality reduction on physical space,''
\newblock {\em IEEE Transactions on Image Processing}, vol. 21, no. 7, pp.
  3239--3252, 2012.

\bibitem{Haupt:2014:onebit}
J.~D. Haupt, N.~D. Sidiropoulos, and G.~B. Giannakis,
\newblock ``Sparse dictionary learning from 1-bit data,''
\newblock in {\em 2014 IEEE International Conference on Acoustics, Speech and
  Signal Processing (ICASSP)}, May 2014, pp. 7664--7668.

\bibitem{Gribonval:10:Identification}
R.~Gribonval and K.~Schnass,
\newblock ``Dictionary identification -- {S}parse matrix-factorization via
  $l_1$minimization,''
\newblock {\em IEEE Transactions on Information Theory}, vol. 56, no. 7, pp.
  3523--3539, 2010.

\bibitem{Aharon:06:Uniqueness}
M.~Aharon, M.~Elad, and A.~M. Bruckstein,
\newblock ``On the uniqueness of overcomplete dictionaries, and a practical way
  to retrieve them,''
\newblock {\em Linear algebra and its applications}, vol. 416, no. 1, pp.
  48--67, 2006.

\bibitem{Spielman:13:Exact}
D.~A. Spielman, H.~Wang, and J.~Wright,
\newblock ``Exact recovery of sparsely-used dictionaries,''
\newblock in {\em Proceedings of the Twenty-Third international joint
  conference on Artificial Intelligence}, 2013, pp. 3087--3090.

\bibitem{Barron:08:MDL}
A.~R. Barron, C.~Huang, J.~Q. Li, and X.~Luo,
\newblock ``The {MDL} principle, penalized likelihoods, and statistical risk,''
\newblock {\em Festschrift for Jorma Rissanen. Tampere University Press,
  Tampere, Finland}, 2008.

\bibitem{Agarwal:2010:flda}
Deepak Agarwal and Bee-Chung Chen,
\newblock ``flda: matrix factorization through latent dirichlet allocation,''
\newblock in {\em Proceedings of the third ACM international conference on Web
  search and data mining}. ACM, 2010, pp. 91--100.

\bibitem{Rendle:2010:FM}
Steffen Rendle,
\newblock ``Factorization machines,''
\newblock in {\em 2010 IEEE International Conference on Data Mining}. IEEE,
  2010, pp. 995--1000.

\bibitem{Chiang:2015:NIMC}
Kai-Yang Chiang, Cho-Jui Hsieh, and Inderjit~S Dhillon,
\newblock ``Matrix completion with noisy side information,''
\newblock in {\em Advances in Neural Information Processing Systems}, 2015, pp.
  3447--3455.

\bibitem{Shin:2015:tumblr}
Donghyuk Shin, Suleyman Cetintas, Kuang-Chih Lee, and Inderjit~S Dhillon,
\newblock ``Tumblr blog recommendation with boosted inductive matrix
  completion,''
\newblock in {\em Proceedings of the 24th ACM International on Conference on
  Information and Knowledge Management}. ACM, 2015, pp. 203--212.

\bibitem{Si:2016:goaldirec}
Si~Si, Kai-Yang Chiang, Cho-Jui Hsieh, Nikhil Rao, and Inderjit~S. Dhillon,
\newblock ``Goal-directed inductive matrix completion,''
\newblock in {\em ACM SIGKDD International Conference on Knowledge Discovery
  and Data Mining (KDD)}, aug 2016.

\bibitem{Negahban:12}
S.~Negahban and M.~J. Wainwright,
\newblock ``Restricted strong convexity and weighted matrix completion:
  {O}ptimal bounds with noise,''
\newblock {\em The Journal of Machine Learning Research}, vol. 13, no. 1, pp.
  1665--1697, 2012.

\bibitem{Cover:12}
T.~M. Cover and J.~A. Thomas,
\newblock {\em Elements of information theory},
\newblock John Wiley \& Sons, 2006.

\end{thebibliography}

\end{document}